\documentclass{article}

\usepackage[preprint,nonatbib]{neurips_2020}

\usepackage[utf8]{inputenc} % allow utf-8 input
\usepackage[T1]{fontenc}    % use 8-bit T1 fonts
\usepackage{hyperref}       % hyperlinks
\usepackage{url}            % simple URL typesetting
\usepackage{booktabs}       % professional-quality tables
\usepackage{amsfonts}       % blackboard math symbols
\usepackage{nicefrac}       % compact symbols for 1/2, etc.
\usepackage{microtype}      % microtypography

\usepackage{graphicx}
\usepackage{subfigure}
\usepackage{multirow}
\usepackage{amssymb}
\usepackage{bm}
\usepackage{color}

\def\D{\mathcal{D}} %domain
\def\E{\bm{E}} 
\def\e{\bm{e}} 
\def\f{\bm{f}} 
\def\F{\bm{F}}

\def\s{\bm{s}}

\def\x{\bm{x}}
\def\v{\bm{v}}

\def\FM{\bm{FM}}
\def\setR{\mathbb{R}}
\def\model{DIFM}

\title{Modeling the Field Value Variations and Field Interactions Simultaneously for Fraud Detection}

\author{
Dongbo Xi$^{1,2,3}$, 
Bowen Song$^{3}$, 
Fuzhen Zhuang$^{1,2,}$\thanks{Corresponding author: Fuzhen Zhuang.} , 
Yongchun Zhu$^{1,2,3}$,\\
\textbf{Shuai Chen}$^{3}$\textbf{,} 
\textbf{Tianyi Zhang}$^{3}$\textbf{,} 
\textbf{Yuan Qi}$^{3}$\textbf{,} 
\textbf{Qing He}$^{1,2}$
\\
$^1$Key Lab of Intelligent Information Processing of Chinese Academy of Sciences (CAS),\\
Institute of Computing Technology, CAS, Beijing 100190, China\\
$^2$University of Chinese Academy of Sciences, Beijing 100049, China\\
$^3$Alipay (Hangzhou) Information \& Technology Co., Ltd.\\
\texttt{\{xidongbo17s,zhuangfuzhen,zhuyongchun18s,heqing\}@ict.ac.cn}\\
\texttt{\{bowen.sbw,shuai.cs,zty113091,yuan.qi\}@antfin.com}
}

\begin{document}

\maketitle
\begin{abstract}
With the explosive growth of e-commerce, online transaction fraud has become one of the biggest challenges for e-commerce platforms. The historical behaviors of users provide rich information for digging into the users' fraud risk. While considerable efforts have been made in this direction, a long-standing challenge is how to effectively exploit internal user information and provide explainable prediction results. In fact, the value variations of same field from different events and the interactions of different fields inside one event have proven to be strong indicators for fraudulent behaviors. In this paper, we propose the Dual Importance-aware Factorization Machines (\model), which exploits the internal field information among users’ behavior sequence from dual perspectives, i.e., field value variations and field interactions simultaneously for fraud detection. The proposed model is deployed in the risk management system of one of the world’s largest e-commerce platforms, which utilize it to provide real-time transaction fraud detection. Experimental results on real industrial data from different regions in the platform clearly demonstrate that our model achieves significant improvements compared with various state-of-the-art baseline models. Moreover, the \model~could also give an insight into the explanation of the prediction results from dual perspectives.
\end{abstract}

\section{Introduction}
With the rapid development of e-commerce and e-payment, the problem of online transaction fraud has become increasingly prominent \cite{cao2019titant}.
For example, according to the statistics\footnote{https://www.chinainternetwatch.com/29999/double-11-2019/}, Alibaba reached single day Gross Merchandise Volume (GMV) of 268 billion CNY in Double 11 Global Shopping Festival \footnote{https://en.wikipedia.org/wiki/Singles\%27\_Day}. A very small portion of fraudulent transaction could easily lead to huge financial loss and threaten the e-commerce business.
%\footnote{https://en.wikipedia.org/wiki/Singles\%27\_Day}
% As an international Fintech company, whose e-commerce platform receives thousands of potentially fraud every day, such deliberate abuse of the e-payment could lead to heavy financial loss and bring great challenge to the e-commerce platform.

\begin{figure*}[!t]
\begin{center}
\subfigure[]{ 
    \label{fig:example}
    \includegraphics[width=.48\columnwidth]{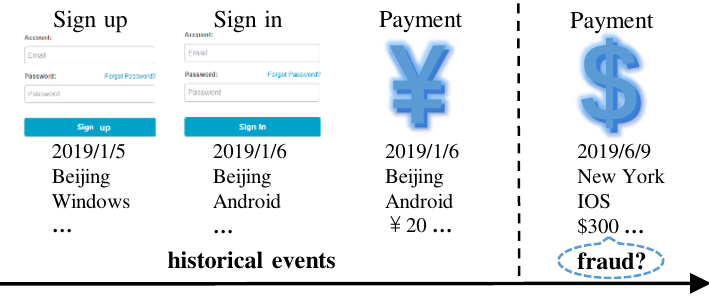} 
  } 
  \subfigure[]{ 
    \label{sample_example} 
    \includegraphics[width=.48\columnwidth]{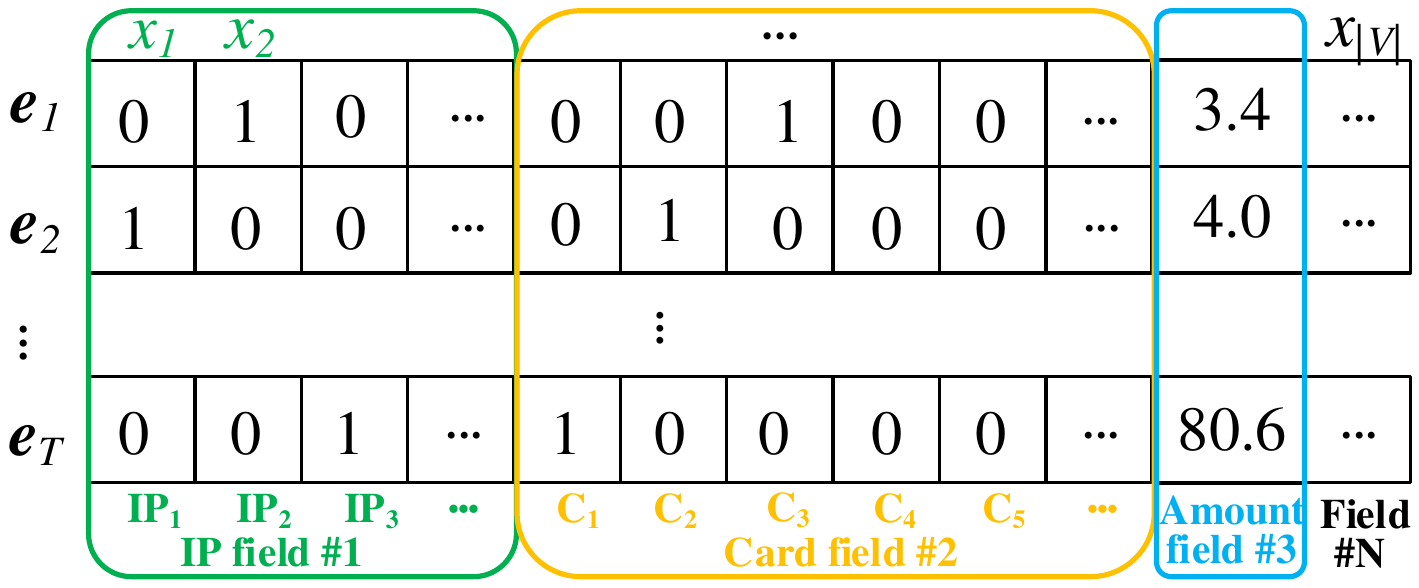} 
  } 
\caption{(a) A typical fraud detection task which exploits the user's historical operation events information for determining the fraudulent risk of target payment event. (b) An example of operation events set $\E$.}
\end{center}
\end{figure*}

Therefore, real-time fraud detection has become a key factor in determining the success of e-commerce platforms. Recently, considerable efforts have shifted from rule-based expert system 
% \cite{cohen1995fast,brause1999neural,rosset1999discovery,baulier2000automated,stefano2001insurance,pathak2005fuzzy} 
\cite{brause1999neural,rosset1999discovery,stefano2001insurance}
to neural network-based models \cite{fu2016credit,wang2017session,zhang2018sequential,jurgovsky2018sequence,cao2019titant,liang2019uncovering} for fraud detection.
The historical events (behaviors) of users provide rich information for digging into the users' fraud risk as shown in Figure \ref{fig:example}.
However, due to model structure limitation, it is difficult for the above methods to exploit the internal field information thoroughly among events (e.g., the field value variations among the history events or field interactions inside each event) nor provide explainable prediction results, which are crucial in fraud detection.

% \begin{figure}[!t]
% 	\begin{center}
% 		\includegraphics[width=0.8\linewidth]{example}
% 		\vspace{-0.3cm}
% 		\caption{The fraud detection task which exploits the historical events of users to help the detection of the target payment event.}
% 		\label{fig:example}
% 	\end{center}\vspace{-0.5cm}
% \end{figure}

% \begin{figure}[t!]
% 	\begin{center}
% 		\includegraphics[width=0.75\linewidth]{event_example}\vspace{-0.2cm}
% 		\caption{A simple example of operation events $\E$.}
% 		\label{sample_example}
% 	\end{center} \vspace{-0.5cm}
% \end{figure}

In this paper, we propose the \textit{Dual Importance-aware Factorization Machines} (\model) to efficiently take advantage of
internal field information among events, which could not only improve the performance of fraud detection but also provide explainable prediction results. \model~captures internal field information from dual perspectives:
1) The \textbf{Field Value Variations Perspective} captures the value variations of each field  between any two events, and the Field Importance-aware module perceives the importance of different field value variations.
%, e.g., IP address changing tends to indicate higher risk than stable one or amount changing.
2) The \textbf{Field Interactions Perspective} models the interactions between all fields inside each event, and the Event Importance-aware module perceives the importance of different historical events.
%, e.g., in card-stolen fraud detection scenario, intuitively, user’s card-binding event is more important than the sign in event.
Besides, the ``wide" layer of \model~can help to score the values of different fields, which could directly indicate the risk level and therefore be used as blacklist/whitelist to screen out the fraudulent/good users.

To summarize, the contributions of this paper are listed as follows:
\begin{itemize}
\item  The \model~effectively and sufficiently exploits internal field information among events from the field value variations and field interactions perspectives simultaneously.
\item The \model~perceives the importance of different field value variations and different events from dual perspectives simultaneously, which can provide explainable prediction results.
\item Experimental results on real industrial data from different regions clearly demonstrate that the proposed \model~model obtains significant improvement comparing to existing state-of-the-art approaches.
\end{itemize}

\section{Related Work}
In this section, we present the related work in two-fold: feature interactions and sequence prediction, fraud detection.

Simply using raw features only could rarely yield optimal results in prediction tasks, therefore more information need to be mined. One way is to learn feature interactions from raw data to generate efficient feature representations. Data scientists made a lot of effort to derive efficient feature interactions from raw data (also known as feature engineering) to obtain better prediction performance \cite{widedeep,lian2017restaurant}.
Instead of generating feature interactions manually, solution \cite{wang2017deepcross} has been proposed to learn feature interactions automatically from the raw data.
Factorization Machine (FM) \cite{rendle2010factorization} is a widely used method to model second-order feature interactions automatically via the inner product of raw embedding vectors. 
Other efforts have also been made to combine the advantages of FM on modeling second-order feature interactions and neural network on modeling higher-order feature interactions 
% \cite{zhang2016fnn,qu2016pnn,he2017nfm,xiao2017afm,guo2017deepfm,lian2018xdeepfm}.
\cite{he2017nfm,xiao2017afm,guo2017deepfm,lian2018xdeepfm}.
However, due to model structure limitation, it is difficult for these methods to exploit the historical information thoroughly. Most of them only simply capture the interactions among all features, which are redundant to the task, since it does not differentiate field information between different events. They do not consider field-in-event-relations and field-between-event-relations, which we believe is important for fraud detection.
The other methods have also attempted to take the user's historical event information into consideration 
% \cite{hidasi2015gru4rec,quadrana2017personalizing,tang2018caser,kang2018self,beutel2018lcrnn,ma2019hierarchical,rakkappan2019context,chen2019behavior,zhou2019dien,tang2019m3r}.
\cite{hidasi2015gru4rec,quadrana2017personalizing,tang2018caser,beutel2018lcrnn,rakkappan2019context,chen2019behavior,zhou2019dien,tang2019m3r}.
Nevertheless, most of these studies mainly focused on the event sequence but ignored the internal field information of historical events.

Fraud detection is a key technology in e-commerce platforms, early researchers mainly focused on rule-based expert system 
% \cite{cohen1995fast,brause1999neural,rosset1999discovery,baulier2000automated,stefano2001insurance,pathak2005fuzzy}.
\cite{brause1999neural,rosset1999discovery,stefano2001insurance}.
These fraud scenarios contain credit card fraud \cite{brause1999neural}, telephone fraud \cite{rosset1999discovery}, insurance fraud \cite{stefano2001insurance} and so on.
With the rapid evolution of fraud patterns, only human-summarized rules or expert knowledge are not sufficient to meet the demand of today's real-time fraud detection \cite{cao2019titant}, therefore some researchers attempted to use machine-learning based methods \cite{tian2015crowd,tseng2015fraudetector,fu2016credit,wang2017session,zhang2018sequential,jurgovsky2018sequence,cao2019titant,liang2019uncovering} to detect fraud.
% For example, graphs have been applied for spotting frauds \cite{tian2015crowd,tseng2015fraudetector,liang2019uncovering}.
Fu et al. (2016) focused on Convolutional Neural Network (CNN) for credit card fraud detection \cite{fu2016credit}.
Other works utilized Recurrent Neural Networks (RNN) for sequence-based fraud detection \cite{wang2017session,jurgovsky2018sequence,zhang2018sequential}.
Liang et al. (2019) utilized Graph Neural Network (GNN) to spot frauds \cite{liang2019uncovering}.
% However, most of these methods are lack of the explainability, which are crucial in fraud detection, while our \model~is able to give reasonable explanations for the prediction results.
% In addition, we try to exploit the internal field information among events from dual perspectives.
However, most of these methods suffer from the same problem: lack of explainability, which is crucial for fraud detection tasks. In this paper, we propose \model~model, which could not only exploit the internal field information among events from dual perspective more thoroughly but also give insight into the explanation of the prediction results.
\section{Methodology}
In this section, we first formulate the problem, then present the details of the proposed model \model.
\subsection{Problem Statement}
A simple example of a user's operation events $\E$ is shown in Figure \ref{sample_example}.
Given a user's operation events $\E = [\e_1, \e_2,..., \e_T]$, where $T$ is the number of the events.
The fields in each events are fixed to collect operation information from a specific aspect, e.g. IP city field or payment amount field, and the number of fields for each event is $N$.
Each field may have multiple values (e.g., in Figure \ref{sample_example}, the IP field has the values of IP$_1$, IP$_2$, IP$_3$ and so on).
$\e_t = [x_1^t, x_2^t,..., x_{|V|}^t]$ $(1\leq t\leq T)$ is the $t$-th event of the user,
where $|V|$ is the number of all different field values (i.e., the dictionary size of field values). 
For categorical fields (e.g., the IP field), $x_i^t$ $(1\leq i\leq |V|)$ is $1$ if $\e_t$ has the value in the current categorical field, otherwise is $0$. For numerical fields (e.g., the Amount field), $x_i^t$ is the real value.
The task is to make a prediction for the current payment event $\e_T$ according to the user's historical operation events $[\e_1, \e_2,..., \e_{T-1}]$ and available information of the current payment event $\e_T$. 
The task can be formulated as Classification, Regression or Pairwise Ranking via utilizing different activation and loss functions according to real application scenarios as described in \cite{rendle2010factorization}.
\subsection{Factorization Machines and Importance-aware Module}\label{FMIN}
In this subsection, we first describe two basic modules of the \model, which are the Factorization Machines and Importance-aware Module, respectively.
\subsubsection{Factorization Machines} 
As shown in Figure \ref{sample_example}, there are rich features in each event which describe the event details in real application scenarios. Since most of the features are one-hot encoding categorical features, the dimension is usually high and the vectors are sparse.
FM is an effective method to address such high-dimension and sparse problems and it can be seen as performing the latent relation modeling between any two field values,
e.g., the field value variations or field interactions.

Firstly, we project each non-zero field value $x_i$ to a low dimension dense vector representation $\v_i$.
Embedding layer is a popular solution in the neural network over various application scenarios.
It learns one embedding vector $\v_i\in\setR^{k}$ ($1\leq i\leq |V|$) for each field value $x_i$.
where $k$ is the dimension of embedding vectors. 
For both categorical and numerical features, we rescale the look-up table embedding via $x_i\v_i$ as done in \cite{he2017nfm}. Therefore, we only need to include the non-zero features, i.e., $x_i\neq 0$.

% Next, in order to extract internal feature interactions of each event for event representation, we apply Hadamard product based FM for $t$-th event $\e^t$ as follow:
Different from traditional FM, which uses inner product to get a scalar,
in neural network, we need to get a vector representation via the Hadamard product as done in \cite{he2017nfm}:
\begin{eqnarray}
FM(\x) = \sum_{i\neq j} x_i\v_i\odot x_j\v_j.\label{equ:fm}
\end{eqnarray}

Hadamard product $\odot$ denotes the element-wise product of two vectors:
$
(\v_i\odot\v_j)_k=v_{ik}v_{jk}.
$
The computing complexity of the above Equation (\ref{equ:fm}) is $O(k|V|^2)$, since all pairwise relations need to be computed.
Actually, the Hadamard product based FM can be reformulated to linear runtime $O(k|V|)$ \cite{he2017nfm} just like the inner product based FM \cite{rendle2010factorization}:
\begin{eqnarray}
FM(\x) = \frac{1}{2}\left[(\sum_{i} x_i\v_i)^2- \sum_{i} (x_i\v_i)^2\right],
\end{eqnarray}
where $\v^2$ denotes $\v\odot\v$.
Besides, in sparse settings, the sums only need to be computed over the non-zero pairs $x_i x_j$. Therefore, the actual computing complexity is $O(k\hat{|V|})$, where $\hat{|V|}$ is the number of non-zero entries in $\x$.
\subsubsection{Importance-aware Module}
In fraud detection, different fields and events often play different roles, which indicate different importance for the detection task. For field importance, IP address changing or amount changing over a short period tends to indicate higher risk than value-stable field, therefore we should pay more attention to IP/amount field if they have such pattern; for event importance, if a user's event has multiple abnormal field values, the event is more important than other normal events.

In order to perceive the relative importance of different fields and events, we design an Importance-aware Module.
We hope the model could pay more attention to important fields and events.
Attention mechanism has been proved to be effective in machine translation \cite{bahdanau2014attention}, where the attention makes the model focus on useful features for the current task.
Inspired by the great success of self-attention in machine translation \cite{vaswani2017selfattention}, 
we design the self-attention-like Importance-aware Module to learn the importance of different fields and events. 
The key advantage of the Importance-aware Module is that it can perceive the importance of different fields and events for each user.
For a vector set $\FM=\{\FM_1,...,\FM_m,...\}$ of different fields or events captured with the FM, the importance weight is defined as scaled dot-product:
\begin{eqnarray}
\hat{a}_m&=&\frac{<F_1(\FM_m),F_2(\FM_m)>}{\sqrt{k}},\\
a_m&=&\frac{\exp(\hat{a}_m)}{\sum_{m}\exp(\hat{a}_m)}, \label{equ:event_weight}
\end{eqnarray}
and the output of the Importance-aware Module ($IM$) is represented as:
\begin{eqnarray}
IM(\FM)=\sum_{m}a_mF_3(\FM_m),\label{IN}
\end{eqnarray}
where $<,>$ represents the dot-product and $F_1$, $F_2$, and $F_3$ represent the feed-forward networks to learn for projecting the input vector to one new vector representation.
It is worth noting that we find using different feed-forward
networks in the designed attention module is more effective than using a single feed-forward network only as adopted in \cite{xiao2017afm, zhou2018din}.

% \begin{figure}[!t]
% \begin{center}
% \includegraphics[width=0.75\linewidth]{model}\vspace{-0.2cm}
% \caption{The proposed \model~model, for simplicity, we do not represent the ``wide" part.}
% \label{fig:model}%
% \end{center}\vspace{-0.5cm}
% \end{figure}%

% \begin{figure}[t!]
% 	\begin{center}
% 		\includegraphics[width=0.75\linewidth]{case2}\vspace{-0.3cm}
% 		\caption{The extracted high-risk fields and events via the Importance-aware Module in two fraud samples.}
% 		\label{fig:case2}
% 	\end{center}\vspace{-0.5cm}
% \end{figure}

\begin{figure*}[!t]
\begin{center}
\subfigure[]{ 
    \label{fig:model}
    \includegraphics[width=.48\columnwidth]{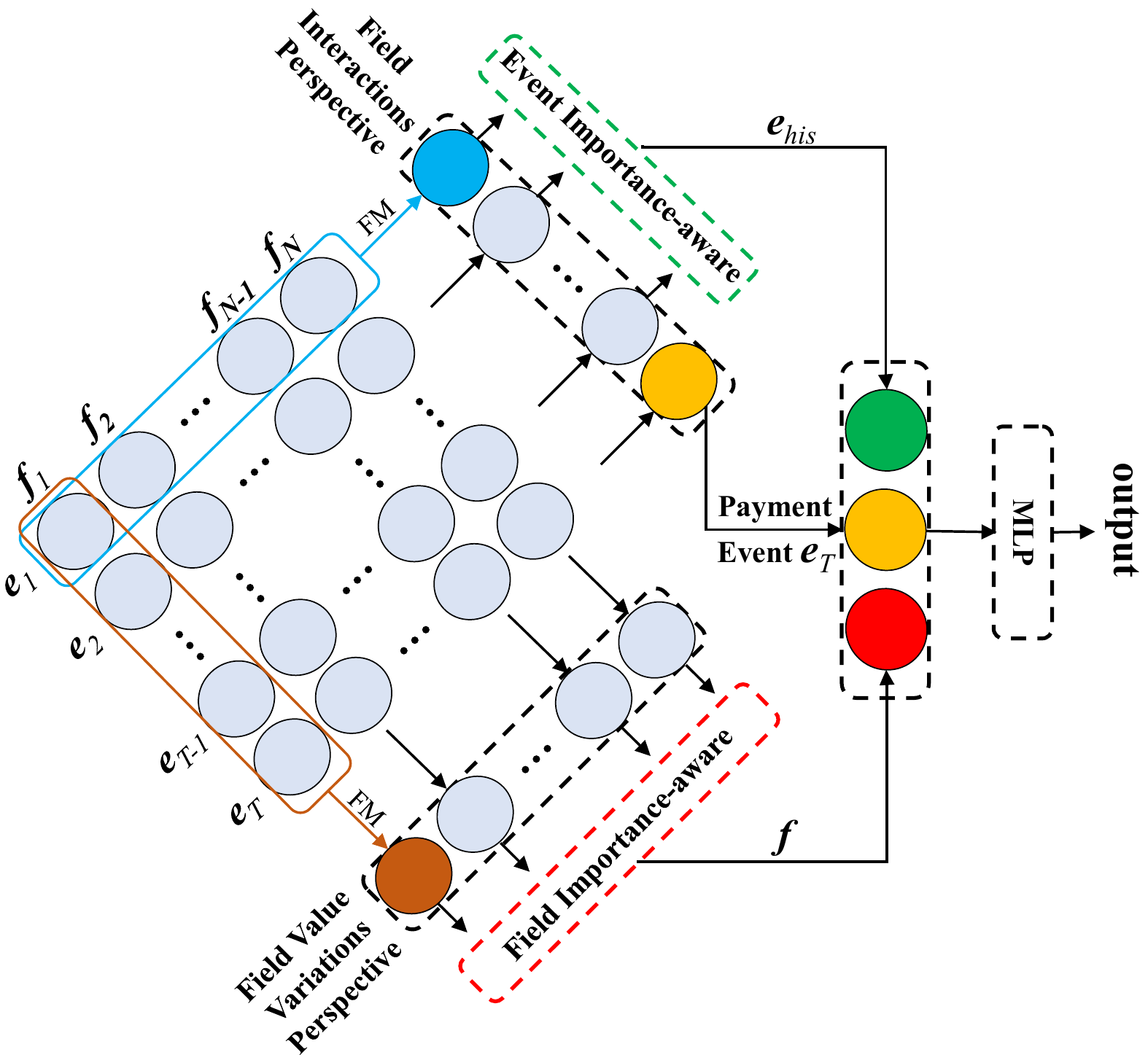} 
  } 
  \subfigure[]{ 
    \label{fig:case2} 
    \includegraphics[width=.48\columnwidth]{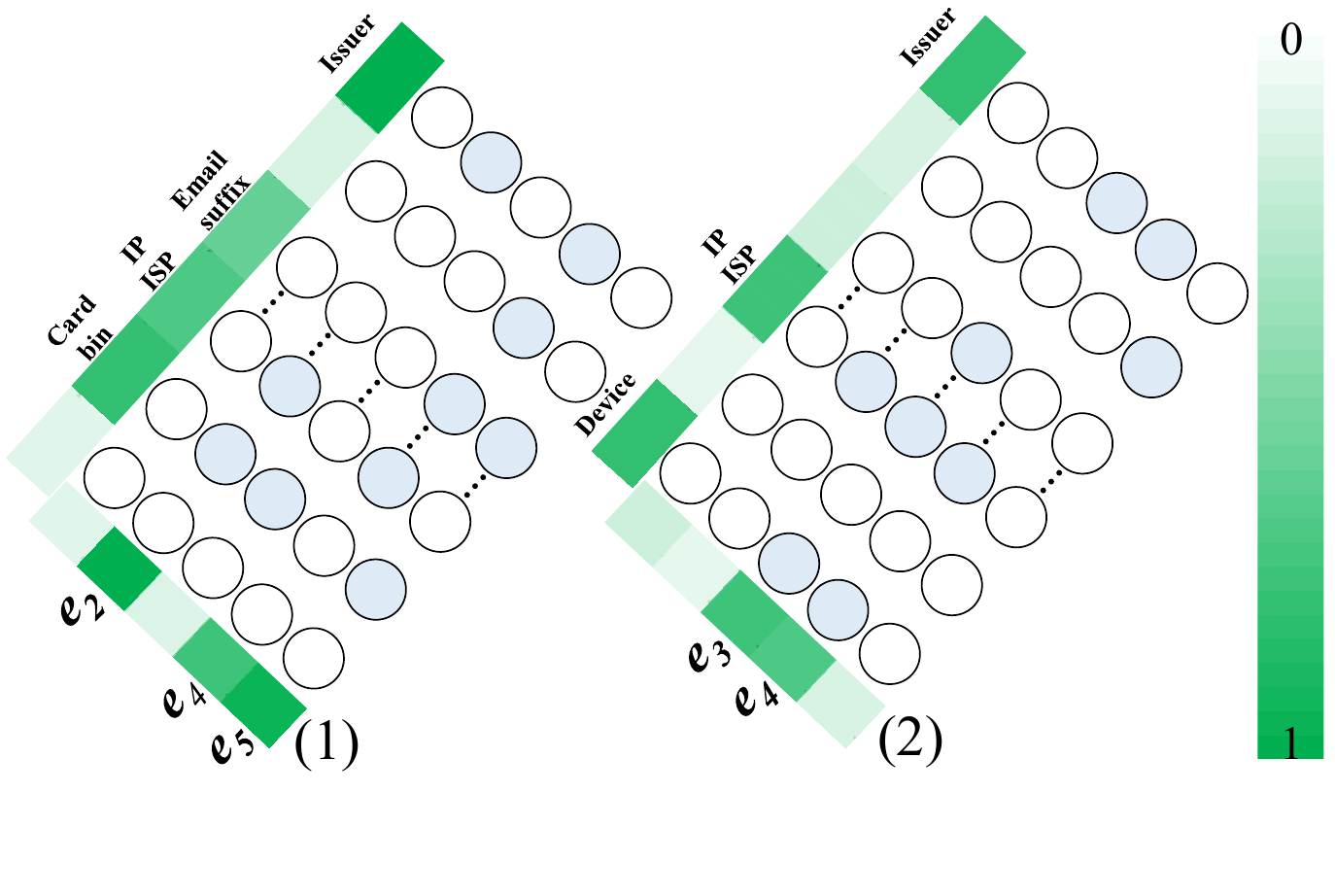} 
  } 
\caption{(a) The proposed \model~model, for simplicity, we do not represent the ``wide" part. (b) The extracted high-risk fields and events via the Importance-aware Module in two fraud samples.}
\end{center}
\end{figure*}

\subsection{Field Value Variations Perspective}
In this subsection, we model the field value variations with the above two modules.

For hacked account, it would be super-expensive to mimic the real account owner’s complete environment information, so field value variations (e.g., the IP changing) in different events tends to indicate higher risk than stable field. 
% For example, in card-stolen fraud detection scenario, 
% the fraudster's behaviors are associated with his/her abnormally frequent payment attempts and the changing of payment environment.

As mentioned in Subsection \ref{FMIN}, FM can help to capture information of the field value variations. Now, we apply the field value variations module FM.
For the $n$-th field $\f_n$, we perform the field value variations in all $T$ events as the brown box shown in Figure \ref{fig:model}:
\begin{eqnarray}
\f_n = FM(\x^n) = \sum_{i=1}^{T-1} \sum_{j=i+1}^{T} x_i^n\v_i^n\odot x_j^n\v_j^n.\label{equ:fm1}
\end{eqnarray}
Now, we apply the field value variations FM (i.e., Equation (\ref{equ:fm1})) to each field in a user's operation events, and we get the representations of all fields $\F = [\f_1, \f_2,..., \f_N]$ thereafter for the Field Importance-aware Module. 
% To the best of our knowledge, this is the first work for fraud detection to consider the field changevalue  among different events.

In fraud detection, if a user's IP address changes over a short period, the field value variation tends to indicate a higher risk than stable fields.
In order to perceive the relative importance of different fields,
for the vector set $\F = [\f_1, \f_2,..., \f_N]$ of different fields captured with the FM, we apply the Field Importance-aware Module in Equation (\ref{IN}) as follows:
\begin{eqnarray}
\f=IM(\F)=\sum_{n=1}^{N}a_nF_3(\f_n),\label{IN1}
\end{eqnarray}
where $a_n$ is the importance weight learnt according to the Equation (\ref{equ:event_weight}).
We hope the model could pay more attention to important fields and focus on useful features for the current task.
As long as a user's field value variations relate to the fraud label, it will be captured by our model via the field value variations perspective.
\subsection{Field Interactions Perspective}
In this subsection, we capture the field interactions with the above two modules.

In fraud detection, a user's fraud behavior can often be detected by multiple features together, since it is easier to spot than via a single feature.
As mentioned in Subsection \ref{FMIN}, FM can be seen as performing the field interactions between any two field values.
%%%%%%%%%%%%%%%%%%%%%%%%%%%%%%%%%%%%%%%%
However, simple calculation of interactions among all features is inefficient and will introduce noise to the model, since the interactions of different fields between different events provide little info for the prediction. For example, fraudster logged in with normal $IP\#1$ in $event\#1$ and buy high-risk item of category $C\#2$ with high-risk $IP\#2$ in $event\#2$, the interaction between features $IP\#1$ and $C\#2$ provides little information, while the internal interaction in $event\#2$ (e.g., interaction between $C\#2$ and $IP\#2$) can capture the event representation better than any single features, and it will improve the model accuracy. Therefore, we capture the field interactions inside each event.
%%%%%%%%%%%%%%%%%%%%%%%%%%%%%%%
For the $t$-th event $\e_t$, we perform the field interactions in all fields as the blue box shown in Figure \ref{fig:model} as follows:
\begin{eqnarray}
\e_t = FM(\x^t) = \sum_{i=1}^{|V|-1} \sum_{j=i+1}^{|V|} x_i^t\v_i^t\odot x_j^t\v_j^t.\label{equ:fm2}
\end{eqnarray}
Thus, we can obtain an effective event representation. 
Some existing methods \cite{wang2017session,zhang2018sequential,tang2019m3r} can also obtain an event representation via using a simple dense layer and the embedding concatenation, but they can not effectively
extract the internal information of each behavior event. 
Now, we apply the field interactions FM (i.e., Equation (\ref{equ:fm2})) to each event in a user's operation behaviors, and we get the representations of all events $\E = [\e_1, \e_2,..., \e_T]$ thereafter for the Event Importance-aware Module. Among them, $\e_T$ is the user's current payment event, whose risk we are trying to model.

Besides, the user's final behavior is strongly correlated with the user's past several behaviors and different historical events of users have different importance.
For example, in card-stolen fraud detection scenario, 
if a user's event has multiple abnormal field values, the event indicates a higher risk than the normal event.
Therefore, a good model could pay more attention to more important historical events.
In order to perceive the relative importance of different events,
for the vector set $\E_{his} = [\e_1, \e_2,..., \e_{T-1}]$ of different historical events captured with the FM, we apply the Event Importance-aware Module in Equation (\ref{IN}) as follows:
\begin{eqnarray}
\e_{his}=IM(\E_{his})=\sum_{t=1}^{T-1}a_tF_3(\e_t),\label{IN2}
\end{eqnarray}
where $a_t$ is the importance weight learnt according to the Equation (\ref{equ:event_weight}).

Thus, the proposed \model~can effectively exploit internal features and perceive the importance of different field value variations and different events from dual perspectives simultaneously. Moreover, the Field and Event Importance-aware Module can provide explainable prediction results  by indicating which field or event is important for the risk score.

The final \model~architecture is shown in Figure \ref{fig:model},
we combine the field value variations feature $\f$ in Equation (\ref{IN1}), the field interactions feature $\e_{his}$ in Equation (\ref{IN2}) and the current prediction event $\e_T$ to feed them to an MLP. 
The output of the MLP is combined with a ``wide" part (for simplicity, we do not represent this part in Figure \ref{fig:model}), and then fed to the $sigmoid$ activation function to form the final \model, which seamlessly combines the field value variations and field interactions perspectives:
\begin{eqnarray}
\s&=&[\f;\e_{his};\e_T],\\
\hat{y}&=&sigmoid(MLP(\s)+f(\x)),
\end{eqnarray}
where $f(\x)$ is the ``wide" part just like the part in Wide\&Deep \cite{widedeep} and $\hat{y}$ indicates the probability of fraud.
The $f(\x)$ is defined as follows:
\begin{eqnarray}
f(\x)=\sum_{t=1}^T\sum_{i=1}^{|V|}w_ix_i^t+w_0, \label{equ:feature_weight}
\end{eqnarray}
where $w_i$ scores the importance of the field value $x_i$, which can indicate high-risk or low-risk for directly using in blacklist or whitelist.

For classification tasks, we need to minimize the \textit{cross entropy} loss:
\begin{equation}
L(\theta)=-\frac{1}{S}\sum^S_{(\x,y)\in\D}(\left(y\log\hat{y}+(1-y)\log(1-\hat{y})\right),
\end{equation}
where $S$ is the number of samples, $y$ is the label of sample $\x$ and $\theta$ is the parameter set, which contains the embedding vector $\v_i$, the weight $w_i$ in the ``wide" part, and the parameters in $F1, F2, F3$ and MLP.

The model is implemented using Tensorflow and training is done through stochastic gradient descent over shuffled mini-batches with the Adam \cite{kinga2015method} update rule.

\section{Experiments}
In this section, we perform experiments to evaluate the proposed \model~model against state-of-the-art methods on real industrial datasets. 
% We first introduce the datasets, baseline methods, implementation details and evaluation metrics of our experiments. Finally, we present our experimental results and further analysis.
\subsection{Datasets}
\begin{table}[!t]
  \centering
  \caption{Summary statistics for the datasets.}
   \setlength{\tabcolsep}{2mm}{
    \begin{tabular}{ccccc}
    \toprule
         Dataset & \#pos & \#neg & \#fields & \#events \\
    \midrule
          C1    & 15K   & 1.37M & 56    & 4.28M \\
          C2    & 10K   & 1.93M & 56    & 3.57M \\
          C3    & 5.7K  & 174K  & 56    & 353K \\
    \bottomrule
    \end{tabular}%
    }
  \label{tab:dataset}
\end{table}%
The statistics of all datasets are shown in Table \ref{tab:dataset}.
The datasets contain the card (debit card or credit card) transaction samples from one international e-commerce platform, which utilizes a risk management system to detect the real-time transaction frauds, e.g., card-stolen cases.
We utilize user's historical events of the last month in three different regions, i.e. countries from Southeast Asia (\textbf{C1},\textbf{ C2}, \textbf{C3}), which consist of 6 event types (i.e., sign up, sign in, digital goods payment, regular goods payment, information modification and card binding). 
And for all events, the following fields are used in our analysis: IP-information, shipping-information, billing-information, card-information, item-category, operation-result, user-account-information, device-information, etc.
The task is to detect whether the current payment event is a card-stolen case.
The fraud labels are from the chargeback from card issuer banks (e.g., the card issuer receives claims on unauthorized charges from the cardholders and reports related transaction frauds to the merchants) and label propagation (e.g., the device and card information are also utilized to mark similar transactions).
\subsection{Baselines}
We compare the proposed method with the following competitive and mainstream models which contain feature interactions based models (W\&D, DeepFM, NFM, AFM, xDeepFM) and event sequence based models (LSTM4FD, LCRNN, M3). For all feature interactions based models, we use all the features of the user's events as the input:
\begin{itemize}
\item \textbf{W\&D}~\cite{widedeep}: It consists of ``wide" and ``Deep" parts, where the ``wide" part is a linear model and the ``Deep" part is an MLP.
\item \textbf{DeepFM}~\cite{guo2017deepfm}: FM and MLP are combined in this model and fed to the output layer in parallel.
\item \textbf{NFM}~\cite{he2017nfm}: This is a simple and efficient neural factorization machine model whose FM is fed to MLP for capturing higher-order feature interactions.
\item \textbf{AFM}~\cite{xiao2017afm}: It adds the attention mechanism to the FM to consider the importance of different pairs. 
\item \textbf{xDeepFM}~\cite{lian2018xdeepfm}: This is the state-of-the-art feature interactions based model which attempts to learn higher-order feature interactions explicitly.
\item \textbf{LSTM4FD}~\cite{wang2017session,zhang2018sequential}: These works have applied LSTM for fraud detection task, and we call these methods as LSTM4FD.
\item \textbf{LCRNN}~\cite{beutel2018lcrnn}: It uses ``Latent Cross" to incorporate contextual data in the RNN by embedding the context feature first and then performing an element-wise product of the context embedding with the model’s hidden states.
\item \textbf{M3}~\cite{tang2019m3r}: This is the state-of-the-art event sequence based model which deals with both short-term and long-term dependencies with mixture models, we choose the better one mixture model M3R-TSL.
% \item \textbf{DIFM-same}: The proposed model utilizes the same $F1, F2, F3$ in the Importance-aware Module.
% \item \textbf{DIFM-$\alpha$}: The proposed model utilizes only the field value change perspective.
% \item \textbf{DIFM-$\beta$}: The proposed model utilizes only the field interaction perspective.
\end{itemize}
\subsection{Implementation Details}
For all datasets, we use: embedding dimension $k$ of 64, the maximum number of events $T$ of 20, one hidden layer of MLP and the dimension is 64, mini-batch size of 256 and learning rate of 0.0005. 
We also use L2 regularization with $\lambda=1e-6$, and dropout  probability is 0.2.
All these values and hyper-parameters of all baselines are chosen via a grid search on the $C1$ validation set. 
We do not perform any datasets-specific tuning except early stopping on validation sets.
The proposed model is trained offline and regularly updated. Meanwhile, the prediction phase is relatively fast, which can meet the requirement of real-time solutions.
% We conduct experiments of all models with NVIDIA GeForce RTX 2080 GPU.
% %and we have anonymously released the source code at https://github.com/xxxxxx. 
\subsection{Evaluation Metrics}
% To evaluate the performance of our proposed \model model and the baselines described above, we follow the existing works \cite{guo2017deepfm,lian2018xdeepfm} to use several standard metrics: \textbf{AUC} (Area Under ROC) and \textbf{LogLoss} (cross entropy).
In binary classification tasks, AUC (Area Under ROC) is a widely used metric \cite{fawcett2006introduction}.
In our real card-stolen fraud detection scenario, 
we should increase the recall rate of the fraudulent transactions, at the same time, disturbing as few normal users as possible.
In other words, we need to improve the True Positive Rate (TPR) on the basis of low False Positive Rate (FPR).
Therefore, we adopt the standardized partial AUC (\textbf{AUC$\bm{_{FPR\leq maxfpr}}$}) \cite{mcclish1989auc} (The standardized area of the head of the ROC curve when the $FPR\leq maxfpr$).
In practice, we require FPR to be less than $1\%$. Hence, we use \textbf{AUC$\bm{_{FPR\leq 1\%}}$} for all experiments.
% Besides, we also focus on some specific points on the head of ROC curve, i.e., the TPR when the FPR are $0.05\%$, $0.1\%$, $0.5\%$ and $1\%$, respectively.
For all experiments, we report the metric with $95\%$ confidence intervals on five runs.
\subsection{Experimental Results}
\begin{table}[t!]
  \centering
  \caption{AUC$_{FPR\leq1\%}$ performance (mean$\pm$95\% confidence intervals) on three datasets.}
  \resizebox{0.62\linewidth}{0.17\linewidth}{
    \begin{tabular}{c||c||c||c}
    \toprule
    \multirow{2}[2]{*}{Model} & C1  & C2  & C3 \\
        & AUC$_{FPR\leq 1\%}$ & AUC$_{FPR\leq 1\%}$ & AUC$_{FPR\leq 1\%}$ \\
    \midrule
    W\&D & 0.6997$\pm$0.0014 & 0.7770$\pm$0.0023 & 0.8202$\pm$0.0092 \\
    DeepFM & 0.7070$\pm$0.0060 & 0.7725$\pm$0.0035 & 0.8476$\pm$0.0070 \\
    NFM & 0.7467$\pm$0.0032 & 0.7931$\pm$0.0074 & 0.8310$\pm$0.0221 \\
    AFM & 0.7086$\pm$0.0047 & 0.7801$\pm$0.0065 & 0.8499$\pm$0.0073 \\
    xDeepFM & 0.7391$\pm$0.0044 & 0.7833$\pm$0.0084 & 0.8557$\pm$0.0137 \\
    \midrule
    LSTM4FD & 0.7118$\pm$0.0087 & 0.7356$\pm$0.0084 & 0.7762$\pm$0.0086 \\
    LCRNN & 0.7189$\pm$0.0068 & 0.7847$\pm$0.0109 & 0.8162$\pm$0.0191 \\
    M3 & 0.7294$\pm$0.0060 & 0.7897$\pm$0.0053 & 0.7618$\pm$0.0268 \\
    \midrule
    DIFM-same & 0.7512$\pm$0.0043 & 0.8275$\pm$0.0063 & 0.8617$\pm$0.0147 \\
    DIFM-$\alpha$ & 0.7401$\pm$0.0070 & 0.8075$\pm$0.0065 & 0.8535$\pm$0.0185 \\
    DIFM-$\beta$ & 0.7640$\pm$0.0105 & 0.8228$\pm$0.0086 & 0.8630$\pm$0.0060 \\
    DIFM & \textbf{0.7681$\pm$0.0088} & \textbf{0.8491$\pm$0.0083} & \textbf{0.8866$\pm$0.0111} \\
    \bottomrule
    \end{tabular}
  \label{tab:result1}%
  }
\end{table}%

% \begin{figure*}[!t]
% 	\begin{center}
% 		\includegraphics[width=0.95\linewidth]{auc_head}
% 		\caption{Mean TPR (when the FPR are $0.05\%$, $0.1\%$, $0.5\%$ and $1\%$, respectively) performance over five runs on datasets of C1, C2 and C3, and the short black lines represent 95\% confidence intervals.}
% 		\label{fig:auc_head}
% 	\end{center}
% \end{figure*}

The experimental results evaluated by AUC$_{FPR\leq1\%}$ on C1, C2 and C3 are presented in Table \ref{tab:result1}. 
We can clearly observe that the performance of baseline models varies on different countries. 

For the C1 dataset, the performance of W\&D is inferior compared with other baselines on AUC$_{FPR\leq1\%}$, 
perhaps because W\&D can not automatically learn feature interactions, and it is just a simple linear model in its ``wide" part.
Besides, DeepFM and AFM obtain similar performance compared with W\&D. DeepFM and W\&D are both parallel structures, although the FM in DeepFM captures second-order feature interactions, it is directly fed to the output layer failing to capture higher-order feature interactions.
Pairs weighted AFM is also directly fed to the output layer, which is lack of higher-order feature interactions.
On the contrary, the FM in NFM is fed to MLP for capturing higher-order feature interactions, so NFM almost performs the best among all baseline models.
Similarly, xDeepFM attempts to learn higher-order feature interactions explicitly, so xDeepFM improves the results compared with W\&D and DeepFM.
For event sequence based models, LSTM4FD and LCRNN obtain similar performance.
M3 obtains further performance improvement due to dealing with both short-term and long-term sequence dependencies with mixture models. However, these models do not consider feature interactions, so the improvement is limited. 

Similar results can also be observed on the C2 dataset for feature interactions based models.
For event sequence based models, the performance of LSTM4FD, which simply apply LSTM for fraud detection, is not good enough.
LCRNN and M3 obtain similar performance and LCRNN incorporating the contextual data in the RNN is effective for the C2 dataset.

For the C3 dataset, it has fewer positive samples, negative samples and available events than C1 and C2 as shown in Table \ref{tab:dataset}.
Therefore, the AUC$_{FPR\leq1\%}$ performance of these models varies a lot, and the $95\%$ confidence intervals of AUC$_{FPR\leq1\%}$ of these baseline models is obviously larger than the performance on C1 and C2 datasets.
Besides, for event sequence based models, LSTM4FD, LCRNN, and M3 are undesirable due to the fewer number of historical events on the C3 dataset.

Besides, the \model~which uses different feed-forward networks in the Importance-aware Module is more effective than the \model-same which only uses a single feed-forward network. 
Moreover, for understanding the contribution of different perspectives in our model,
we construct ablation experiments over DIFM-$\alpha$, DIFM-$\beta$ and \model.
DIFM-$\alpha$ and DIFM-$\beta$  utilize the field value variations perspective and field interactions perspective in their models, respectively.
The field interactions perspective brings greater gain than the field value variations perspective.
Furthermore, the proposed \model~model can efficiently take advantage of
internal field features among events from dual perspectives and obtain the best performance on all three datasets, which indicates both two perspectives matter for the performance.
Instead of capturing sequence info directly, our \model~capture more fine-grained value-variations information. Frequent value-variations of certain field in event sequences is a signal which strongly correlates to fraudulent activities. For example, user with IP changing from $IP\#1$ to $IP\#2$ indicates higher risk comparing to user with stable IP. \model~captures value-variations of each field between any two events in user’s operation sequences via the Field Value Variations module. Combined with Field Interactions module, which captures field-in-event-relations, our proposed framework captures useful info more effectively. That's why it performs better than the RNN-based baselines.
These improvements also indicate that the proposed \model~model can better handle the fraud detection task.
% Similar conclusions can also be observed in the experimental results evaluated by TPR (when the FPR are $0.05\%$, $0.1\%$, $0.5\%$ and $1\%$, respectively) on datasets of C1, C2 and C3, which are presented in Figure \ref{fig:auc_head}.
\begin{table}[!t]
  \centering
  \caption{The extracted high risk (high weight) and low risk (low weight) field values according to the learned weights in Equation (\ref{equ:feature_weight}), the ``Card bin" is the last six digits of the card number, the ``IP ISP" is Internet Service Provider for the IP, the ``Email suffix" is the suffix of the email and the ``Issuer" is the name of the issuing bank.}
  \resizebox{0.65\linewidth}{0.13\linewidth}{
    \begin{tabular}{cllll}
    \toprule
        & Card bin & IP ISP & Email suffix & Issuer \\
    \midrule
     & \#1 (94/105) & \#1 (5/8) & \#1 (28/28) & \#1 (76/79) \\
     High   & \#2 (46/46) & \#2 (30/30) & \#2 (59/59) & \#2 (136/137) \\
    Risk    & \#3 (12/12) & \#3 (22/22) & \#3 (56/60) & \#3 (105/113) \\
        & \#4 (78/82) & \#4 (4/5) & \#4 (149/183) & \#4 (77/85) \\
    \midrule
     & \#5 (0/58) & \#5 (0/81) & \#5 (0/1120) & \#5 (1/1008) \\
     Low   & \#6 (0/181) & \#6 (0/384) & \#6 (0/101) & \#6 (0/34) \\
    Risk    & \#7 (0/38) & \#7 (0/89) & \#7 (3/5958) & \#7 (0/20) \\
        & \#8 (0/239) & \#8 (0/38) & \#8 (0/79) & \#8 (0/54) \\
    \bottomrule
    \end{tabular}%
    }
  \label{tab:case1}%
\end{table}%
\subsection{Case Study}
In this subsection, we make some analysis on the explainability of the proposed \model~on C3 dataset.

Firstly, we extract four highest-risk and lowest-risk field values from four fields according to the learned weights of ``wide" part in Equation (\ref{equ:feature_weight}).
We present the field values in Table \ref{tab:case1}.
To conform to Data-Protection-Regulation of the company, we replace the real field values with \#number and present the ratio of each field value with (fraud number/total number).
We can clearly observe that these extracted field values are very strong indicators of fraud or not.
For example, the Email suffix \#2 occurs 59 times and all of them are fraud samples, while the Email suffix \#5 occurs 1120 times and all of them are normal usage.
Therefore, these field values can be directly added to the blacklist and whitelist.
By the way, in our risk management system, model predication is used together with rule-based methods, and this gives double insurance to our system.

Then, we extract some high-risk fields and events for two fraud samples according to the learned importance weights in Equation (\ref{equ:event_weight}). 
We normalize the weights.
The field value variations of each field and field interactions of each event can be regarded as modeling of users' fraud patterns.
We present the fraud patterns in Figure \ref{fig:case2}.
The solid circle represents that the field value has changed since the last event.
The depth of color for each square illustrates the distributions of importance weights. The darker the color is, the higher weight it has.

The fields of Card bin, IP ISP, Email suffix and Issuer should be relatively stable for normal users, but for sample (1), the values of these fields change multiple times in the account's operation history, which indicates a high risk.
Therefore, these fields obtain higher weights and stable fields have lower weights.
Meanwhile, the events 2, 4, and 5 have multiple abnormal field values, so they are more important than other normal events. 
The distribution of the event weights also confirms this.
The similar pattern can also be observed in the sample (2).

These above observations demonstrate that our \model~model can effectively find the important fields and events from dual perspectives.
These above results also demonstrate that the proposed \model~has the capability to provide explainable prediction results.
These explanations are of great help to the experts in fraud detection for further analysis.
\section{Conclusion}
In this paper, we designed a \model~model for online transaction fraud detection. 
Specifically, \model~ utilizes the Factorization Machines and the proposed Importance-aware Module to exploit the internal field information among behaviors from dual perspectives, which are field value variations perspective and field interactions perspective.
The extensive experimental results on real-world datasets from a world-leading e-commerce platform clearly demonstrate the performance improvements of our proposed model compared with various state-of-the-art baseline methods and the case study further approves the explainability of our model.

\bibliographystyle{named}
\bibliography{neurips_2020}

% \renewcommand*{\bibfont}{\small}
% \printbibliography

\end{document}